\documentclass[runningheads]{llncs}
\usepackage{amsmath,amsfonts}%
\usepackage{bbm}
\usepackage{bm}
\usepackage{graphicx}%
\usepackage{multirow}%
\usepackage{xcolor}%
\usepackage{textcomp}%
\usepackage{algorithm}%
\usepackage{algorithmicx}%
\usepackage{algpseudocode}%

\begin{document}
\def\bbx{\mathbf x} 
\def\bby{\mathbf y} 
\def\bbz{\mathbf z} 
\newcommand{\bfm}[1]{\ensuremath{\mathbf{#1}}}

\newcommand{\red}{\textcolor{red}}
\newcommand{\blue}{\textcolor{blue}}

\def\bx{\bfm x}   \def\bX{\bfm X}  \def\XX{\mathbb{X}} 
\def\bY{\bfm Y}   \def\YY{\mathbb{Y}}
\def\bz{\bfm z}   \def\bZ{\bfm Z}  \def\ZZ{\mathbb{Z}}

\def\RR{\mathbb R}

\newcommand{\bfsym}[1]{\ensuremath{\boldsymbol{#1}}}
\def \bsbeta{\bfsym \beta}
\def \bstheta{\bfsym \theta}
\def\bfh{\boldsymbol h} 
\def \bfy{\mathbf y}

\def\bfh{\boldsymbol h} 
\def\boldq{\boldsymbol q}
\def\boldp{\boldsymbol p}
\def\boldz{\boldsymbol z}

\def\calA{{\cal  A}} \def\cA{{\cal  A}}
\def\calB{{\cal  B}} \def\cB{{\cal  B}}
\def\calC{{\cal  C}} \def\cC{{\cal  C}}
\def\calD{{\cal  D}} \def\cD{{\cal  D}}

\def \bsbeta{\bfsym \beta}
\def \bstheta{\bfsym \theta}
\def\ba{\bfm a}   \def\bA{\bfm A}  \def\AA{\mathbb{A}}

\def\PP{\mathbb{P}}

\def \CE {\text{CE}}

\title{Knowledge Distillation of LLMs for Automatic Scoring of Science Assessments}
% \author{Author Names Omitted for Anonmity}

\author{Ehsan Latif\inst{1,2} \and
Luyang Fang\inst{1,3} \and
Ping Ma \inst{3}\textsuperscript{,*} \and
Xiaoming Zhai\inst{1,2}\textsuperscript{,*}
}
\institute{AI4STEM Education Center, Athens, GA, USA \and
Department of Mathematics, Science, and Social Studies Education, University of Georgia, Athens, GA, USA \and
Department of Statistics, University of Georgia, Athens, GA, USA \\
*\email{\{pingma,xiaoming.zhai\}@uga.edu}}

\maketitle              % typeset the header of the contribution
\begin{abstract}
This study proposes a method for knowledge distillation (KD) of fine-tuned Large Language Models (LLMs) into smaller, more efficient, and accurate neural networks. We specifically target the challenge of deploying these models on resource-constrained devices. Our methodology involves training the smaller student model (Neural Network) using the prediction probabilities (as soft labels) of the LLM, which serves as a teacher model. This is achieved through a specialized loss function tailored to learn from the LLM's output probabilities, ensuring that the student model closely mimics the teacher's performance. To validate the performance of the KD approach, we utilized a large dataset, 7T, containing 6,684 student-written responses to science questions and three mathematical reasoning datasets with student-written responses graded by human experts. We compared accuracy with state-of-the-art (SOTA) distilled models, TinyBERT, and artificial neural network (ANN) models. Results have shown that the KD approach has 3\% and 2\% higher scoring accuracy than ANN and TinyBERT, respectively, and comparable accuracy to the teacher model. Furthermore, the student model size is 0.03M, 4,000 times smaller in parameters and x10 faster in inferencing than the teacher model and TinyBERT, respectively. The significance of this research lies in its potential to make advanced AI technologies accessible in typical educational settings, particularly for automatic scoring.

\end{abstract}

\section{Introduction}
\label{introduction}
Artificial Intelligence (AI) in education has evolved from a theoretical concept to a practical tool, significantly impacting classroom assessment practices and adaptive learning systems \cite{gonzalez2021artificial,holmes2022state}. AI for personalized learning and assessment provides opportunities for more tailored and effective educational experiences \cite{zhai2020applying}.
Integrating Large Language Models (LLMs) from domains on AI like BERT \cite{devlin2018bert} into education has been a significant milestone in enhancing learning experiences, providing personalized learning content and support, and facilitating automatic scoring  \cite{liu2023context,zhai2022chatgpt,selwyn2019should,zhai2021review}. Despite their potential, the deployment of LLMs in educational settings is constrained by their considerable size (714MB for 178 million parameters and 495MB for 124 million parameters) and computational requirements (16 Tensor Processing Units), presenting a challenge for widespread adoption in resource-constrained educational environments such as mobiles/tablets and school-provided laptops with no GPUs or TPUs and limited memory \cite{hinton2015distilling}.

To bridge this gap, our study explores the feasibility of distilling the knowledge of LLMs into smaller neural networks, referred to as \textit{student models}, with fewer parameters and hidden layers. By training a smaller student model using soft labels provided by a fine-tuned LLM (i.e., \textit{teacher model}), we aim to achieve a similar scoring performance to that of LLMs, but with reduced model size.
% This approach leverages recent advancements in knowledge distillation (KD) \cite{hinton2015distilling} and AI, demonstrating the potential of smaller models to achieve comparable accuracy to LLMs \cite{zhang2022cross,li2021data,xu2017cross}.

% In this context, we have fine-tuned BERT on a large dataset of student-written used by Liu et al. \cite{liu2023context} science assessment responses. The fine-tuned model then serves as a teacher model, guiding the training of a compact student model. Our innovative loss function is designed to align the student model's predictions with the teacher model's, achieving similar accuracy with a smaller model size and faster inference time. This technique is particularly applicable for automatic scoring (evaluation of student written resposnes) in education, where timely and accurate feedback is essential \cite{zhai2022applying,zhai2021meta}.

The significance of this research lies in its potential to make advanced AI technologies accessible in typical educational settings. The study addresses the technical challenges of deploying AI models in resource-constrained environments and highlights the potential of AI to transform educational assessment practices. By enabling the deployment of efficient automatic scoring systems on less powerful hardware available in school settings, we contribute to the democratization of AI in education.
% This approach aligns with the growing demand for personalized learning experiences and adaptive assessment tools in the educational sector to address educational equity and diversity \cite{zhai2020applying, yuan2021research, bhunia2021text, adhikari2020exploring}.
%This paper presents a method for distilling the knowledge of a fine-tuned LLM into a smaller neural network for educational purposes, particularly automatic scoring. 
The key contributions of this paper are:
\vspace{-5pt}
\begin{itemize}
    \item  We demonstrate the successful application of a novel knowledge distillation (KD) strategy that, while inspired by \cite{hinton2015distilling}, is uniquely adapted and optimized for the context of educational content. 
    % This includes the development of a specialized loss function designed to maximize the fidelity of the student model to the teacher model's output probabilities, specifically tailored for the nuances of language and conceptual knowledge in science assessments.
    \item Our approach achieves a significant reduction in model size and computational requirements without compromising accuracy. The student model, distilled from a fine-tuned BERT teacher model, exhibits a model size that is 4,000 times smaller and demonstrates an inference speed that is ten times faster than that of its teacher counterpart.
    \item Through comprehensive evaluations using a large dataset of 10k student-written responses to science questions, our work not only validates the effectiveness of our KD method against state-of-the-art models like TinyBERT \cite{jiao2019tinybert} and generic ANN models \cite{ghiassi2012automated}, but also highlights its superior performance. 
    % \item Our research offers valuable insights into the practical application of knowledge distillation in education technology. By analyzing the specific challenges and requirements of deploying AI models in resource-constrained educational environments, we provide a roadmap for future research and development in this area.

\end{itemize}

\section{Proposed Knowledge Distillation}

KD is a technique to transfer knowledge from a trained large model (teacher) to a more compact and deployable model (student). We take inspiration from the prominent KD approach, introduced by \cite{hinton2015distilling}, which involves using the class probabilities generated by the pre-trained large model as \textit{soft labels} for training the smaller model, effectively transferring its predictive and generalization capabilities.
Building on this concept, we develop a method for applying KD in the context of automated scoring systems, aiming to improve the process of evaluating educational content using AI.

\iffalse
\begin{algorithm}
\caption{Knowledge Distillation for Education Technology}
\begin{algorithmic}[1]
\Require Training dataset $\mathcal{D}$, Teacher model $T$, Learning rate $\eta$, Regularization parameter $\lambda$, Number of training epochs $E$
\Ensure Trained Student model $S$

\State Initialize student model $S$ with parameters $\bstheta$

\For{epoch $= 1, \ldots, E$}
    \For{each batch $\mathcal{B} \subseteq \mathcal{D}$}
        \State Initialize gradient $\nabla_{\bstheta} \gets 0$
        
        \For{each data point $(\bx_i, y_i) \in \mathcal{B}$}
            \State Compute teacher prediction: $\boldp_i \gets T(\bx_i)$
            \State Compute student prediction: $\boldq_i \gets S(\bx_i, \bstheta)$
            \State Compute loss:
            \State $\quad \mathcal{L}_{\text{KD}} \gets \CE(\boldp_i, \boldq_i) + \lambda \cdot \CE(y_i, \boldq_i)$
            \State Accumulate gradients: $\nabla_{\bstheta} \gets \nabla_{\bstheta} + \frac{\partial \mathcal{L}_{\text{KD}}}{\partial \bstheta}$
        \EndFor
        
        \State Update student model parameters:
        \State $\quad \bstheta \gets \bstheta - \eta \cdot \nabla_{\bstheta}$
    \EndFor
\EndFor

\State \Return $S$
\end{algorithmic}
\label{algorithm}
\end{algorithm}
\fi

Specifically, for each data point $\bx_i$ in the training sample $\mathcal{D}$, the teacher model predicts the class probability $\boldp_{i}=(p_{i1},\ldots , p_{iK})^T$, where $p_{ij}$ represents the predicted probability that the $i^{\text{th}}$ data point belongs to class $j$. 
The student model is trained using both the training sample $\mathcal{D}$ and the corresponding \textit{soft labels} $\boldp = \{\boldp_{i}\}_{i=1}^N$ produced by the teacher model.
% As formulated in Section \ref{sec:NN},
We represent the student model by a neural network $f(\cdot, \bm{\theta})$. The discrepancy between the student and teacher models is measured as
\begin{small}
\begin{equation}\label{eq:student_teach_discrepancy}
\begin{aligned}
\tilde{\mathcal{L}}( f(\cdot,\bstheta); \mathcal{D}, \boldp ) &= \frac{1}{N}\sum_{i=1}^N \CE(\boldp_i, f(\bx_i,\bstheta)), \\
 &= - \frac{1}{N}\sum_{i=1}^N\sum_{k=1}^K p_{ik} \log\left( f_k(\bx_i;\bstheta) \right),
 \end{aligned}
\end{equation}
\end{small}
\noindent which is the sample mean of the cross-entropy $\CE(\boldp_i, f(\bx_i,\bstheta))$ across $i$. 
To leverage the information from both the training data  and the teacher model's predictions, KD aims to solve 
\begin{small}
\begin{equation}\label{eq:opt_KD_loss}
\begin{aligned}
    \bstheta_{\mathrm{KD}}^* 
    &= \underset{\bstheta \in \RR^d}{\arg \min }\left\{\mathcal{L}^{\mathrm{KD}}( f(\cdot , \bstheta); \mathcal{D}, \boldp, \lambda) \right\} \\
    &= \underset{\bstheta \in \RR^d}{\arg \min }\left\{ \mathcal{L}( f(\cdot , \bstheta) ; \mathcal{D} ) + \lambda\tilde{\mathcal{L}}( f(\cdot,\bstheta); \mathcal{D}, \boldp )  \right\},
\end{aligned}
\end{equation}
\end{small}
where the minimized KD loss $\mathcal{L}^{\mathrm{KD}}( f(\cdot , \bstheta); \mathcal{D}, \boldp, \lambda)$ is the linear combination of two loss terms in KD loss and (\ref{eq:student_teach_discrepancy}). 
The first term of the KD loss equation measures the discrepancy between the predictions of the student model and the actual labels. The second term assesses the prediction discrepancy between the student and teacher models. In this context, $\lambda$ serves as a constant that balances the impact of these two aspects of the loss. Setting $\lambda=0$ reduces the KD loss to the conventional empirical risk loss. 
% Setting $\lambda=0$ reduces the KD loss in Eq. (\ref{eq:opt_KD_loss}) to the conventional empirical risk loss. 
% The pseudo code for the proposed KD approach can be seen in Alg.~\ref{algorithm}

%------
\begin{figure}[h]
\vspace{-0.5cm}
\centering
\includegraphics[width=0.75\linewidth]{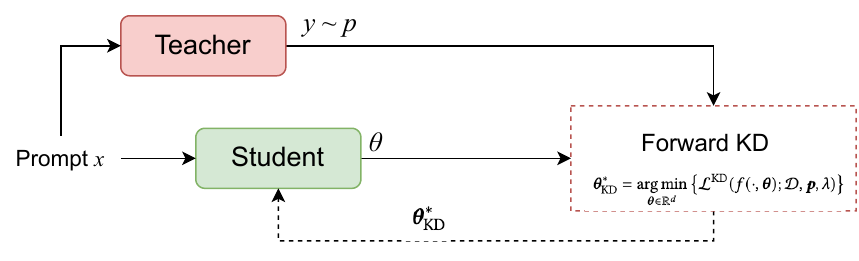}
\vspace{-0.5cm}
\caption{The architecture of the proposed KD approach uses prediction probabilities as soft labels from the teacher model and forces the student model to achieve these prediction probabilities through the fitting loss function.}
% The procedure begins with a pre-trained teacher model from which we derive knowledge from class probabilities. This knowledge, along with the data, is then utilized to inform the training of a student model.}
\label{fig:workflow}
\vspace{-0.6cm}
\end{figure}
%------

KD enables the student model to attain performance comparable to the teacher model while considerably reducing the computational resources required for training.
The teacher model's predicted probability outputs $\boldp$ provide valuable insights into its data interpretation. By minimizing the discrepancy between the outputs of the student and teacher models, the student model can effectively adopt the knowledge and insights of the teacher model. Consequently, despite its simpler architecture and reduced computational resources, the student model can match the performance of the more complex teacher model.

In Fig.~\ref{fig:workflow}, we present the architecture of the proposed KD method. With a well-performing, fine-tuned, large teacher model and a new dataset, we run the teacher model on the dataset and extract knowledge to guide the training of a more compact student model. In this study, we extract the class probabilities predicted by the teacher model as the knowledge to be transferred to the student model. sing both the knowledge from the teacher model and the data from the dataset, we train the student model based on optimization, as outlined in Equation (\ref{eq:opt_KD_loss}).

%------------------------------
%------------------------------
\section{Experimental Setup}\label{experiments}

Our study investigates whether a significantly smaller neural network can effectively mimic the capabilities of a fine-tuned LLM through the proposed KD strategy. Additionally, the study explores how this approach can enhance model performance.
We apply our proposed methodology across diverse datasets to train a compact model to achieve this goal. This model is then compared with the SOTA TinyBERT \cite{jiao2019tinybert} and a trained smaller ANN \cite{ghiassi2012automated} to evaluate the performance in terms of accuracy and efficiency. 
% The rationale behind selecting TinyBERT \cite{jiao2019tinybert}, and ANN \cite{ghiassi2012automated} models for comparison in our study stems from their relevance and distinct characteristics in the context of knowledge distillation and neural network efficiency. 
TinyBERT stands out due to its specific design to compress the size and computational demands of BERT through a sophisticated distillation process, making it an ideal benchmark for assessing the efficiency and effectiveness of our distillation approach. On the other hand, ANN represents a broader category of neural networks that, while not specialized in natural language processing tasks to the same extent as TinyBERT, provides a contrasting baseline to evaluate the general applicability and performance of our distilled model in automatic scoring. The comparison against these models allows us to validate the superiority of our approach not only against a state-of-the-art distilled model like TinyBERT, which is directly relevant to our domain but also against more generic neural network architectures.

\subsection{Data Collection and Preprocessing}
The study utilized a meticulously categorized dataset of student-written responses to a science question and three mathematical assessment items, each falling under the multi-class category for automatic scoring. Each student response in datasets is graded by a human expert for automatic scoring, and human scores are used for validation. On average, each student's written textual response contains 15 words. The detailed composition of each assessment item's dataset is presented in Table \ref{tab:real_des}.

\vspace{-5pt}
\begin{table}[h]
    \centering
    \scalebox{0.9}{
    \begin{tabular}{l|p{2cm}p{1.5 cm}p{4cm}p{3cm}} 
    \hline
     Dataset & Sample size & Classes & Teacher & Student  \\ \hline 
     7T & 6,684 & 10 & SciEdBERT (114M) & E-LSTM (0.03M)  \\
     Bathtub & 1,145 & 5 & BERT\_base (110M)  & E-LSTM (0.03M)  \\
    Falling Weights & 1,148 & 4 & BERT\_base (110M) & E-LSTM (0.03M)  \\
    Gelatin & 1,142 & 5 & BERT\_base (110M)  &  E-LSTM (0.03M) \\
    \hline
    \end{tabular}}
    \vspace{6pt}
    \caption{Sample size and the teacher and student model used for each dataset. The number of parameters for each model is shown in parentheses.}
    \label{tab:real_des}
    \vspace{-15pt}
\end{table}

\subsection{Dataset}
The 7T dataset is a large dataset consisting of seven tasks from the SR1 dataset, including short constructed student responses and human-expert graded scores. Overall, the 7T dataset consists of 6,684 labeled student responses from \cite{zhai2022applying}, similar to the dataset used for SciEdBERT by Liu et al. \cite{liu2023context}. We utilized three multi-class assessment tasks from the Mathematical Thinking in Science (MTS) project \cite{ETS2023} responded to by high-school students: Bathtub, Falling Weights, and Gelatin containing 1,145, 1,148, and 1,142 student-written responses respectively. Each dataset contains a different number of classes (scores assigned by human experts) for student-written responses. More specific details about scoring and assessment items can be found in \cite{latif2024fine}. This comprehensive dataset facilitated a nuanced analysis of the capacity of compact scoring models for student-written responses, ensuring robust and broadly applicable study findings. We processed each dataset by excluding empty responses and ensuring text-formatted student responses and ranged labels.
    % , specifically, (915 for task H4-2, 915 for task H4-3, 834 for task H5-2, 883 for task J2-2, 743 for task J6-2, 739 for tasks J6-3, and 845 for task R1-2). 
%     \item \textbf{Falling Weights:} We also have taken the student response dataset for challenging mathematical thinking assessment item for falling weight similar to the data used by Latif \& Zhai \cite{latif2024fine}, which consists of 1,148 student written and human-expert graded scores.
%     \item \textbf{Gelatin:} Another dataset for gelatin assessment items contains 1,142 student responses, considered samples to train teacher models and distill knowledge to student models.
% \end{itemize}

\begin{table*}[t]
    \centering
    \scalebox{0.92}{
    \begin{tabular}{p{3cm}|p{2cm}p{2.5cm}p{2.5cm}p{2.5cm}} 
    \hline
    & \multicolumn{4}{c}{ Accuracy } \\
     & Teacher & TinyBERT & ANN  & KD  \\ \hline 
    7T & 0.891$\pm$0.016 & 0.752$\pm$0.003 & 0.716$\pm$0.002 &  0.757$\pm$0.001* \\
    Bathtub & 0.938$\pm$0.014 & 0.833$\pm$0.019 & 0.831$\pm$0.021 &   0.852$\pm$0.012* \\
    Falling Weights & 0.904$\pm$0.013 & 0.856$\pm$0.015 & 0.865$\pm$0.014  &  0.888$\pm$0.008* \\
    Gelatin & 0.871$\pm$0.018 & 0.735$\pm$0.010 &  0.739$\pm$0.011 &  0.780$\pm$0.014* \\
    \hline
    & \multicolumn{4}{c}{ F-1 Score } \\ \hline 
    7T & 0.842$\pm$0.017  & 0.749$\pm$0.009 & 0.706$\pm$0.001 & 0.751$\pm$0.005*  \\
    Bathtub & 0.914$\pm$0.021 & 0.832$\pm$0.069 & 0.830$\pm$0.024 & 0.851$\pm$0.011*  \\
    Falling Weights & 0.893$\pm$0.018 & 0.855$\pm$0.015 & 0.864$\pm$0.014 &  0.886$\pm$0.009*  \\
    Gelatin & 0.804$\pm$0.016 & 0.731$\pm$0.008 & 0.733$\pm$0.014 & 0.766$\pm$0.017*  \\
    \hline
    \multicolumn{5}{p{12cm}}{* KD has shown higher accuracy and F-1 score than TinyBERT and ANN, and is comparable to the Teacher model for each dataset.}\\
    \end{tabular}}
    \vspace{2pt}
    \caption{Accuracy and F-1 score performance comparison of teacher, TinyBERT \cite{jiao2019tinybert}, ANN \cite{ghiassi2012automated}, and KD model for benchmark datasets. The mean accuracies and standard deviations are displayed.}
    \label{tab:real_acc}
    \vspace{-0.6cm}
\end{table*}

%---- ----

\subsection{Training Scheme}

\subsubsection{Model Setup}
This study uses SciEdBERT \cite{liu2023context} with 114M parameters as a specialized Science Education BERT model, and the standard BERT base model \cite{devlin2018bert} contains 110M parameters as the teacher model. These models have been shown to perform brilliantly in processing textual data. For performance comparison, we used TinyBERT \cite{jiao2019tinybert} with 67M parameters and small ANN \cite{ghiassi2012automated}. 
For the KD method, we constructed a compact neural network with an embedding layer with an output dimension of 32 and a bidirectional LSTM layer with 16 units (significantly fewer parameters than transformers), followed by a GlobalMaxPooling1D layer. Further, it includes two dense layers, with the first having 16 neurons and 'relu' activation, and the final layer is equipped with a softmax activation for multi-class classification.
Additionally, dropout layers are integrated for regularization, and the model is optimized with Adam.

% \begin{itemize}
%     \item \textbf{Bathhub:} The model features an embedding layer with an output dimension of 32 and a bidirectional LSTM layer with 16 units, followed by a GlobalMaxPooling1D layer. It includes two dense layers, with the first having 16 neurons and 'selu' activation, and the final layer is equipped with a softmax activation for multi-class classification.
%     \item \textbf{7T:}  The model consists of an embedding layer with an output dimension of 64, followed by a single LSTM layer with 64 units. It includes two dense layers, with the first having 32 neurons and 'relu' activation and the final layer equipped for multi-class classification with a softmax activation.
%     \item \textbf{Falling Weights:} The model comprises an embedding layer with an output dimension of 32, a bidirectional LSTM layer with 16 units, a global max pooling layer, and two dense layers, including a final softmax activation for multi-class classification. 
%     \item \textbf{Gelatin:} The model consists of an embedding layer with an output dimension of 128, enhanced with L1 regularization for sparsity. It includes a single LSTM layer with 64 units, followed by a dense layer with 16 neurons, using 'selu' activation and both L1 and L2 regularization. The final layer is a dense layer equipped for multi-class classification, featuring a softmax activation.
% \end{itemize}

% Additionally, dropout layers are integrated for regularization, and the model is optimized with Adam.

%---- ----

\subsubsection{Evaluation and Validation}

We partition each dataset into training, validation, and testing sets in a $7:1:2$ ratio. The model optimization employs cross-entropy loss, and to prevent overfitting, an early stopping callback that monitors the validation loss is utilized.  We present the prediction accuracy on the test set to assess the model's performance.

% \subsection{Dataset}

The summary of the dataset and the teacher and student (KD) models used for each dataset is detailed in Table \ref{tab:real_des}. We provide the number of parameters for each model in parentheses. The student model is much smaller than the teacher model.

\subsection{Results}

%---- ----
%---- ----

The comparative analysis of model accuracy across four datasets is presented in Table \ref{tab:real_acc}. Results reveal the efficacy of KD in enhancing the performance of a student model as compared to the SOTA TinyBERT \cite{jiao2019tinybert}  and ANN \cite{ghiassi2012automated} for text classification, in terms of both accuracy and F-1 score. Furthermore, it also provides close accuracy and F-1 score as the complex teacher model. The Falling Weights dataset serves as a typical example, with KD providing performance comparable to the teacher model, suggesting that even models with much smaller sizes can achieve similar performance to the large teacher model.
% We observe that the accuracy of KD (0.903) is similar to the original neural network model, indicating that the accuracy of automaticscoring can be improved by including the model training with KD strategy.
We observed that KD outperforms TinyBERT and ANN in accuracy by 2.5\% and 3.2\%, respectively, and in F-1 score by 2.2\% and 3.0\%, respectively.
This observation highlights the superiority of KD over SOTA model distillation approaches.
Considering both accuracy, F-1 score (shown in Table~\ref{tab:real_acc}), and model size (shown in Table~\ref{tab:real_des}), results highlight the practicality and applicability of the KD approach for automatic scoring on resource constrained-devices. 
% This improvement delineates the potential of KD to augment model capabilities, particularly in scenarios where the SOTA approaches may not fully capture the underlying patterns in the data.

Despite the success of KD, it is essential to recognize that the student models, although improved, usually do not reach the performance benchmarks set by the teacher models. This is notably apparent in the 7T dataset; the integration of KD leads to better performance compared to the ANN and TinyBERT but still does not match the teacher models' accuracy. Such a discrepancy can be attributed to the inherent limitations of the student models, which possess simpler architectures and are trained on smaller datasets with far fewer training parameters. 
% \textcolor{red}{Update results for Tiny BERT}

% \begin{table*}[t]
%     \centering
%     \scalebox{0.95}{
%     \begin{tabular}{c|p{2.5cm}p{2.5cm}p{2cm}p{2cm}} 
%     \hline
%     & \multicolumn{4}{c}{ Accuracy } \\
%      & Teacher & TinyBERT & ANN  & KD  \\ \hline 
%     Bathtub & 0.886 & 0.852 & 0.824 &   0.856* \\
%     7T & 0.891 & 0.740 & 0.705 (0.017) &  0.756* (0.011) \\
%     Falling Weights & 0.925 & 0.870 & 0.840  &  0.888* \\
%     Gelatin & 0.871 & 0.776 &  0.763 &  0.784* \\
%     \hline
%     \multicolumn{5}{p{8.5cm}}{* KD has shown higher accuracy than TinyBERT and Original NN and is comparable to the Teacher model for each dataset.}\\
%     \end{tabular}}
%     \caption{Accuracy performance comparison of teacher, TinyBERT \cite{jiao2019tinybert}, ANN \cite{ghiassi2012automated}, and KD model for benchmark datasets. }
%     \label{tab:real_acc}
% \end{table*}

% We also observe that the effect of KD compared to that of the ANN strategy varies across datasets. The Gelatin dataset sees the smallest improvement, indicating that the ANN was already quite effective for this dataset. On the other hand, for the other three datasets, KD successfully improves the performance of the student model. Across all datasets, while KD does not achieve the same level of accuracy as the Teacher model, it greatly reduces the performance gap, demonstrating its efficiency in establishing compact student models.

The results demonstrate that the KD strategy is a powerful tool in model training, beneficial for applications such as automatic scoring. By effectively condensing the knowledge of a large, pre-trained model into a more compact one, KD not only improves performance but also facilitates the deployment of such models in resource-constrained environments.

\subsection{Sensitivity Analysis}

We investigated the impact of the hyperparameter \(\lambda\) in Eq. (\ref{eq:opt_KD_loss}) on scoring accuracy to learn more about the resilience of the KD approach. We assessed the KD approach using a step size of 0.02 and a range of \(\lambda\) values from 0.08 to 0.02 for the Bathtub dataset. Although there were little variations in the accuracy of the KD technique with the modification of \(\lambda\), consistently outperformed the baseline models. These findings suggest that the KD approach is comparatively resistant to the selection of \(\lambda = 0.2\), demonstrating consistent performance across a range of hyperparameter values.

\iffalse
%------
\begin{figure}[h]
%\vspace{-0.2cm}
\centering
\includegraphics[width=0.99\linewidth]{sensitivity.png}
%\vspace{-0.2cm}
\caption{ Results of the sensitivity analysis for the hy[erparameter $\lambda$. The pool of possible $\lambda$ is set as $\{0.08, 0.1, 0.12, 0.14, 0.16, 0.18, 0.2\}$. Each point represents the average outcome of three replicates at a specific $\lambda$, with error bars depicting the standard deviation. Red dashed line represents the result using the student model and the blue dashed line represents the result of TinyBERT model.}
\label{fig:sensitivity}
\end{figure}
%------
\fi

%-------------------------------

\section{Discussion}
\label{discussion}
The results of this study highlight the revolutionary possibilities of KD in educational technology, especially in light of the limitations of standard school computing resources. The use of KD in education represents a substantial breakthrough, particularly in automated grading systems. Nevertheless, like any emerging technology, it is important to recognize its limitations as well as its potential for development in the future. In the traditional education system, automatic evaluation is yet a point of discussion  \cite{selwyn2019should}. Therefore, our proposed solution is a supplementary tool designed to support and not replace traditional assessment methods established in the education system. Further studies and educational policy adaptations are necessary to fully integrate such technologies into formal school environments.

% \subsection{Application of KD in Education}
% \label{application_of_kd_in_education}
The most noteworthy application of KD in education is the creation of accurate and efficient automatic scoring systems. A major challenge in many educational contexts is that traditional scoring systems can demand extensive processing resources to function successfully on school-setting devices such as entry-level laptops and tablets. This problem is addressed by KD, which enables the creation of ``student models" with significantly lower processing requirements while preserving much of the accuracy and efficiency of larger ``teacher models.

% When combined with automatic scoring, these distilled models can give students offline fast, reliable, and objective feedback—an essential feature of offline adaptive learning environments without needing high processing and an internet connection. Accurate and timely feedback is crucial for creating a personalized and interesting learning environment, which contemporary educational environments are calling for more and more.

Furthermore, KD models are ideally suited for integration into tablet- and smartphone-based learning apps due to their smaller size and reduced processing requirements. The capacity to run complex AI models on these devices, which are increasingly prevalent in educational contexts, creates new opportunities for interactive and adaptable learning experiences.

\section{Conclusion}
\label{conclusion}
This study effectively illustrates how KD can be used to optimize LLMs for usage in educational technology, especially on low-processor devices. We maintain great accuracy with a much smaller model size (0.03M parameters) and processing requirements by condensing the knowledge of LLMs into smaller neural networks. The distilled models perform better than SOTA TinyBERT and ANN models on various datasets, demonstrating the efficacy of this approach even though their parameter sizes are up to 100 times less than teacher models. This work has important applications since it provides a method to incorporate cutting-edge AI tools into conventional school environments, which frequently have hardware constraints. The learning process and accessibility of personalized education technology can be significantly improved by the capacity to implement effective and precise AI models for uses such as autonomous scoring. Essentially, this work establishes the foundation for future developments in the field and validates the viability of KD in educational contexts, underscoring the significance of ongoing research and innovation in AI for education. In the future, we will work on processing soft-labels and prompt processing to avoid amplification of faults of teacher models by employing more sophisticated techniques.

\section*{Acknowledgment}

This work was funded by the National Science Foundation(NSF) (Award Nos. 2101104, 2138854) and partially supported by the NSF under grants DMS-1903226, DMS-1925066, DMS-2124493, DMS-2311297, DMS-2319279, DMS-2318809, and by the U.S. National Institutes of Health under grant R01GM152814.

\bibliographystyle{splncs04}
\bibliography{references}

\end{document}